\documentclass[11pt,a4paper]{article}
\usepackage[hyperref]{ranlp2023}
\usepackage{times}
\usepackage{latexsym}

\pdfoutput=1

\usepackage{microtype}

\usepackage{pgfplotstable}
\usepackage{pgfplots}
\usepackage{microtype}

\usepackage[utf8]{inputenc} 
\usepackage[T1]{fontenc}    
\usepackage{hyperref}       
\usepackage{url}            
\usepackage{booktabs}       
\usepackage{amsfonts}       
\usepackage{nicefrac}       
\usepackage{microtype}      
\usepackage{lipsum}
\usepackage{graphicx}
\usepackage{xcolor}
\usepackage{pgfplots}
\pgfplotsset{width=7cm,compat=1.9}
\graphicspath{{media/}}

\usepackage{times}
\usepackage{url}
\usepackage{latexsym}
\usepackage{verbatim}
\usepackage{tabularx}
\usepackage{supertabular}
\usepackage{amssymb,amsmath,array}
\usepackage{multirow}
\usepackage{rotating}
\usepackage{subcaption}

\def\arrvline{\hfil\kern\arraycolsep\vline\kern-\arraycolsep\hfilneg}

\def\totallyunseensentences{definitely unseen sentences}

\aclfinalcopy

\title{The Dark Side of the Language:\\ Pre-trained Transformers in the DarkNet}

\author{ Leonardo Ranaldi $^{(*)}$,  Aria Nourbakhsh$^{(*)}$, Arianna Patrizi$^{(*)}$, Elena Sofia Ruzzetti$^{(*)}$, \\
\textbf{Dario Onorati$^{(\bullet)}$, Michele Mastromattei$^{(*)}$, Francesca Fallucchi$^{(*)}$, Fabio Massimo Zanzotto$^{(*)}$ } \\
  $^{(*)}$  
Human-Centric ART Group, Department of Enterprise Engineering, University of Rome Tor Vergata.\\ 
  $^{(\bullet)}$ Department of Computer, Control and Management Engineering,
University of Rome "Sapienza".\\
 {\tt [first name].[last name]@uniroma2.it}  }

\date{}

\begin{document}
\maketitle
\begin{abstract}
Pre-trained Transformers are challenging human performances in many NLP tasks. 
The massive datasets used for pre-training seem to be the key to their success on existing tasks. 
In this paper, we explore how a range of pre-trained Natural Language Understanding models perform on 
\totallyunseensentences{} provided by classification tasks over a DarkNet corpus. Surprisingly, results show that syntactic and lexical neural networks perform on par with pre-trained Transformers even after fine-tuning. Only after what we call extreme domain adaptation, that is, retraining with the masked language model task on all the novel corpus, pre-trained Transformers reach their standard high results. This suggests that huge pre-training corpora may give Transformers unexpected help since they are exposed to many of the possible sentences.
\end{abstract}

\section{Introduction}

Transformers \cite{zhang2019ernie,Radford2018ImprovingLU:GPT} have been rocking the field of NLP. These Transformers are outperforming all previous methods and, sometimes, even humans in many NLP tasks \cite{wang-etal-2018-glue,wang2020superglue,AMMUS-T-PTLMs:2021,ThreatsPTLMs:2022}. 

Pre-training on large corpora seems to be the key that boosts performances, as these Transformers may induce clear models of target languages. 
Indeed, BERT is pre-trained on an English corpus of 3,300M words consisting of books \cite{BookCorpusUsedInBERT} and Wikipedia. The English version of the last ERNIE \cite{Sun2021ERNIE3L} is trained on an even more extensive corpus. MEGATRON-LM \cite{Shoeybi2019MegatronLMTM} utilizes an incredible corpus of 174 GB, and the Chinese version of ERNIE breaks the records by exploiting a 4TB corpus \cite{ERNIE_3.0_4TB}. Therefore, the challenge is training over always more massive corpora. 

Huge pre-training corpora may give unexpected help to Transformers: their successes in downstream tasks can be because Transformers have seen large parts of possible sentences. This could be a sort of overfitting. This possible shortcoming is sometimes considered when novel Transformers are introduced \cite{Radford2019LanguageMA,Shoeybi2019MegatronLMTM}.  For this reason, \citet{Radford2019LanguageMA} have excluded Wikipedia pages for pre-training as it is a common data source for other downstream datasets. Yet, when using off-the-shelf pre-trained models, this caution is generally disregarded. 
For example, the discovering ongoing conversation (DOC) task, introduced by \citet{10.1145/2885501} was found challenging for humans, but the BERT baseline model achieved the astonishing 88.4 F1 score \cite{wang2020superglue}. DOC consists of determining if two utterances are contiguous in classical theatrical plays. These 
plays may be included in the book dataset \cite{BookCorpusUsedInBERT} used for pre-training BERT. This may explain the superhuman performance in such a challenging task.

Corpora and related tasks derived from the DeepWeb and DarkWeb \cite{info13090435,DBLP:conf/itasec/RanaldiNFZ22,avarikioti2018structure,Choshen2019TheLO} offer a tremendous opportunity to study Transformers and other natural language models on 
\totallyunseensentences{}.

Performances on these tasks cannot depend on overfitting over-seen sentences.

Indeed, it is extremely unlikely that texts extracted from these sources are included in pre-training corpora. Moreover, language on the DarkNet may have very different characteristics with respect to the one accessible from the surface web \cite{Choshen2019TheLO}.  

\begin{table*}[h!]
\centering 
\tabcolsep=0.15cm
\begin{tabular}{l|cc|cc|c|c|}
  & \multicolumn{2}{|c|}{Onion Drugs} & \multicolumn{2}{|c|}{Onion Forums} & \multicolumn{2}{|c|}{Surface Web} \\
  & \textbf{Legal} & \textbf{Illegal} & \textbf{Legal} & \textbf{Illegal} & \textbf{eBay drugs} & \textbf{DBpedia sample} \\ 
\hline
\hline
\textbf{\#tokens} & 41,683 & 67,506 & 41,683 & 43,654 & 114,817
 & 46,792\\
\textbf{\#types} & 8,576 & 12,334 & 8,576 & 9,411 & 18,405
 & 7,941\\
\textbf{types/token ratio} & 4,86  & 5,47 & 4,86 & 4,36 & 6,23 & 5,88 \\
\hline
\hline
\textbf{BERT's types pieces/OOVs ratio} & 2,94  & 3,96 & 3,67 & 3,15 & 5,03 & 2,40 \\
\end{tabular}
\caption{Lexical description of the corpus subsets and their lexical coverage with respect to the vocabulary of BERT compared with a sample of the DbPedia dataset \cite{NIPS2015_250cf8b5}.}
\label{tab:covered_token}
\end{table*}

In this paper, we aim to explore how pre-trained Natural Language Understanding models behave on \totallyunseensentences{}. 
These \totallyunseensentences{} are provided by the DarkNet corpus along with a classification task. We experimented with Stylistic Classifiers based on the bleaching text model  \cite{van-der-goot-etal-2018-bleaching}, with Lexical Neural Networks based on GloVe \cite{pennington-etal-2014-glove} and word2vec \cite{mikolov2013efficient}, with Syntatic-based neural networks based on KERMIT \cite{zanzotto-etal-2020-kermit}, and with holistic Transformers such as BERT \cite{devlin-etal-2019-bert}, XLNet \cite{Yang2019XLNetGA}, ERNIE \cite{zhang2019ernie} and Electra \cite{clark2020electra}. Results show that syntactic and lexical neural networks surprisingly outperform pre-trained Transformers even after fine-tuning \cite{FineTuning:Wei2021WhyDP}. Only when pre-training is extended to the novel corpus and Transformers see these \totallyunseensentences{} do their performances increase to the expected level.
This seems to suggest that huge pre-training corpora may give Transformers the unexpected help of showing them many possible sentences.

The rest of the paper is organized as follows: Material and Methods, Results and Discussion, Conclusions, and Limitations.
The code and data are publicly available at:  \url{https://github.com/ART-Group-it/Transformers_DarkWeb}.

\section{Material and Methods}

This section describes the corpus used to challenge Transformers that is a source of \totallyunseensentences{} (Section \ref{sec:dark-dataset}) and the investigated transformers along with other neural network methods (Section \ref{sec:methods}).

\subsection{Material: A Dark Web Dataset}
\label{sec:dark-dataset}

Corpora scraped from DarkWeb to fight illegal actions offer a tremendous opportunity for studying how large pre-trained models behave on \totallyunseensentences{}. Indeed, as discussed in section \ref{sec:methods}, pre-trained Transformers as well as ``pre-trained'' syntactic neural networks have not used any of these corpora for training. Hence, corpora collected for totally different reasons can help to shed light on an important research question: is pre-training on large datasets a sort of overfitting on many of the possible sentences? 

Classifying legal and illegal actions is a key task in the DarkWeb (also called Onion Web). Hence, it is also crucial to try to understand whether or not Transformers or other neural network-based models add value to models for this challenging task.

\subsubsection{Using an Existing Corpus: DUTA-10k}
\label{sec:duta}

For our experiments, we used the ``Darknet Usage Text Addresses'' (DUTA-10k) \citep{Nabki2019ToRankIT} as exploited in \citet{Choshen2019TheLO}. 
The corpus DUTA-10k \citep{Nabki2019ToRankIT} contains onion web data manually tagged in legal and illegal samples. \citet{Choshen2019TheLO} selected only the drug subdomain and used four different subsets of 571 samples each: (1) legal onion drugs, (2) illegal onion drugs, (3) onion forums discussing legal activities, and (4) onion forums discussing illegal topics. Additionally, to compare with the data from surface web, \citet{Choshen2019TheLO} have extracted item descriptions from eBay as well.  These descriptions were selected by searching the keywords, which are, `marijuana', `weed', `grass', and `drug'. 
The resulting DUTA-10K used in \citet{Choshen2019TheLO} and in our experiments contains 5 subsets (see Table \ref{tab:examples} for some examples): the four manually annotated onion web sets and the eBay drugs set.  

\citet{Choshen2019TheLO} propose to use data from DUTA-10k for the task of classifying legal and illegal activities.
Hence, the five subsets are used to produce four different classification datasets: (1)~eBay vs. legal drugs; (2) legal vs. illegal drugs; (3) legal vs. illegal forums; and, finally, (4) legal and illegal drugs as training data vs. legal and illegal forums as testing dataset. The last task is the most important and complex as it tests knowledge transferability as training is on the drug dataset and the testing is on a totally different domain.

\subsubsection{Retrieving DUTA-10K and Final Corpus}
\label{sec:duta_adaptation}
Using the corpus proposed in \citet{Choshen2019TheLO} is not straightforward as DUTA-10k has to be reconstructed by scraping the DarkWeb again. Indeed, for legal reasons, the dataset contains only manually tagged links to the DarkWeb\footnote{Data and code are available in \citet{Choshen2019TheLO}'s GitHub repository \url{https://github.com/huji-nlp/cyber}}. 

In our experiments, we then extracted the corpus and prepared the dataset by using links and the tools provided by \citet{Choshen2019TheLO}. 
The preprocessing consists of removing:
HTML tags, non-linguistic content such as buttons, encryption keys, metadata, and common words such as ``Show more results''.
All the 571 samples for all the subsets still exist in the DarkWeb and, thus, have been used to create our dataset. The resulting subsets (see Tab.~\ref{tab:covered_token}) are not extremely large in terms of tokens.

Since there is not a standard split in the provided datasets, we prepared 5 different 70\%-30\%  splits in training and testing of each subtask. Results presented in \citet{Choshen2019TheLO} are obtained on a single 70\%-30\% split. However, in our experiments, we opted for these 5 different 70\%-30\% splits to evaluate the stability of experimental results. Results may vary from split to split. Indeed, our results differ from the experiments conducted in \citet{Choshen2019TheLO} when we tried to replicate their models.

In addition to the provided datasets, we also used a random subset of the DBpedia dataset \cite{NIPS2015_250cf8b5} for comparing the language of the datasets with a more general language.

\subsection{Methods: Classification Models}
\label{sec:methods}

Our goal is to investigate how Transformers and other pre-trained models perform on \totallyunseensentences{}. This section introduces the models used in this study: lexical-based neural networks (Sec.~\ref{sec:lexical_based_neural_networks}), syntax-based neural networks (Section~\ref{sec:kermit}), and Transformers (Section~\ref{sec:holistic_tranformers}).
The description of the pre-trained models discusses the size of the corpora used to train each model as well.

To discard the idea that 
the chosen task has strong stylistic signals where Transformers perform poorly (see results for the corpus linguistic acceptability task in \citet{warstadt2018neural}), this section utilizes two style-oriented classifiers (Section~\ref{sec:stylistic_classifiers}) to investigate whether determining legal and illegal activities is indeed only a stylistic task. The code can be found in the file \textit{Code The Dark Side of the Language.zip} and will be made publicly available.

\subsubsection{Style-oriented Classifiers}
\label{sec:stylistic_classifiers}
Legal and illegal activities may be described with different styles of language: a formal vs. a more informal style of writing. For this reason, we use two style-oriented classifiers as detectors to understand if this task is merely stylistic. 

The first is a family of classifiers that exploits \emph{part-of-speech (POS) tags}, treating them as bag-of-POSs. These are very simple models which mainly capture the distribution of POSs in target texts. In line with \citet{Choshen2019TheLO}, we tested two non-neural models, namely Naive Bayes NB(POS) and Support Vectors Machines. In addition, we tested BoPOS, a simple feed-forward neural network using bag-of-POS as input representation.

\emph{Bleaching text} \cite{van-der-goot-etal-2018-bleaching} is a model proposed to capture the style of writing. 
Originally, it has been applied to cross-lingual authors' gender prediction. 
This model converts sequences of tokens, e.g., `1x Pcs Mobile Case!? US\$65', into abstract sequences according to these rules presented with the effect on the example:\\ 
\begin{tabular}{p{7cm}}
    - each token is replaced by its length (effect: `02 03 06 06 05')\\
\end{tabular}\\
\begin{tabular}{p{7cm}}
    - alphanumeric characters are merged into one single letter and other characters are kept (effect: `W W W W!? W\$W')\\    
\end{tabular}\\
\begin{tabular}{p{7cm}}
    - punctuation marks are transformed into a unified character (effect: `W W W WPP W')\\
\end{tabular}\\
\begin{tabular}{p{7cm}}
    - upper case letters are replaced with `u', lower case letters with `l', digits with `d', and the rest to `x' (effect: `dl ull ull ullxx uuxdd')\\
\end{tabular}\\
\begin{tabular}{p{7cm}}
    - consonants are replaced with `c', vowels to `v' and the rest to `o' (effect: `oc ccc cvcvcv cvcvoo vcooo')\\
\end{tabular}\\
Finally, a sample is represented by the concatenation of all the above transformations. For classification, we use a linear SVM classifier with a binary bag of word representation.

\subsubsection{Lexical-based Neural Networks}
\label{sec:lexical_based_neural_networks}
To investigate the role of pre-trained word embeddings, we used a classifier based on vanilla feed-forward neural networks (FFNN) over bag-of-word-embedding (BoE) representations. In BoE(GloVe), sentence representations are computed as the sum of word embeddings representing their words.
BoE(GloVe) uses GloVe word embeddings \cite{pennington-etal-2014-glove} trained on 2014 Wikipedia dumps and Giga5 (see Table \ref{tab:pretraining}). 
The supporting FFNN of BoE(GloVe) consists of an input layer of dimension 300 and 2 hidden layers of 150 and 50 dimensions with the $ReLU$ activation function.

\subsubsection{Syntactic-based Neural Networks}
\label{sec:kermit}
To evaluate the role of ``pre-trained'' universal syntactic models, we used the Kernel-inspired Encoder with Recursive Mechanism for Interpretable Trees (KERMIT) \cite{zanzotto-etal-2020-kermit}. This model positively exploits parse trees in neural networks as it increases the performances of pre-trained Transformers when it is used in combined models. 

\begin{table*}[h!]
\centering 
\tabcolsep=0.06cm
\begin{tabular}{l|cccc}
\hline
 & \textbf{eBay/Legal Drugs} & \textbf{Drugs} & \textbf{Forums} & \textbf{Drugs/Forums} \\ 
\hline

\textit{NB (POS)} \cite{Choshen2019TheLO} & $91.4$ & $77.6$  & $74.1$ & $78.4$ \\

\textit{SVM (POS)} \cite{Choshen2019TheLO} & $63.8$ & $63.8$ & $85.3$ & $62.1$ \\ 

\hline

(our) \textit{NB (POS)}  & $90.4(\pm 1.3)$ & $83.7(\pm 0.7)$  & $64.3(\pm 2.2)$ & $48.2(\pm 1.8)$ \\

(our) \textit{SVM (POS)} & $86.6(\pm 0.6)$ & $83.6(\pm 1.2)$ & $61.6(\pm 1.5)$ & $43.2(\pm 2.3)$ \\ 

\hline
\textit{Bleaching text} & $84.73(\pm 0.8)$ & $81.3(\pm 0.6)$  & $56.65(\pm 1.7)$  & $55.68(\pm 1.3)$ \\ 
\hline

\end{tabular}
\caption{Accuracies of style-oriented classifiers over 5 different splits on the Legal vs. Illegal Classification Task and accuracies of NB (POS) and SVM (POS) as reported in \cite{Choshen2019TheLO}}
\label{tab:style}
\end{table*}

The version used in the experiments encodes parse trees in vectors of 4,000 dimensions. The rest of the feed-forward network is composed of 2 hidden layers of dimension 4,000 and 2,000 respectively, and finally the output layer of dimension 2. Between each layer, the $ReLU$ activation function and a dropout of 0.1 is used to avoid overfitting on the train data.

Even in this case, the model is somehow `pre-trained'. In fact, KERMIT exploits parse trees produced by a traditional parser. In our experiments, we used the English constituency-based parser in CoreNLP \cite{zhu-etal-2013-parser-fast}. The parser is trained on the standard WSJ Penn Treebank \cite{marcus-etal-1993-building}, which contains only around 1M words.

\subsubsection{Holistic Transformers}
\label{sec:holistic_tranformers}
We tested the following Transformers to cover the majority of cases of pre-training size (see Table~\ref{tab:pretraining}) and models:

\paragraph{BERT} \cite{devlin-etal-2019-bert}, the architecture Bidirectional Encoder Representations from Transformers, trained on the BooksCorpus \cite{zhu2015aligning} and English Wikipedia ($BERT_{base}$). 

\paragraph{XLNet} \cite{Yang2019XLNetGA}, a generalized autoregressive pre-training technique that allows the learning of bidirectional contexts by maximizing the expected likelihood over all permutations of the factorization order and to its autoregressive formulation. XLNet is trained on 32.89 billion tokens, taken from datasets gathered from the surface web or publicly available datasets, such as Wikipedia, Bookcorpus, Giga5, Clueweb, and Common Crawl.

\paragraph{ERNIE} \cite{Sun2021ERNIE3L}, an improved language model that addresses the inadequacy of BERT and utilizes an external knowledge graph for named entities. ERNIE is pre-trained on Wikipedia corpus and Wikidata knowledge base.
    
\paragraph{ELECTRA} \cite{clark2020electra}, an improved BERT where, instead of masking input tokens, these are ``corrupted'' with replacement tokens that potentially fit the place. The training procedure is a classification of each token on if it is a corrupted input or not. To make its performance comparable to BERT, they have trained the model on the same dataset that BERT was trained on.

The models for all the proposed Transformers were implemented using the Transformers library from Huggingface and the pre-trained version of AutoModelforSequenceClassification \cite{wolf-etal-2020-transformers}. 
For each model, we chose the Adam optimizer \cite{Kingma2015AdamAM} with a batch size of $32$ and fine-tuned the model for 5 epochs, following the original paper \cite{devlin-etal-2019-bert}. For hyperparameter tuning, the best learning rate is different for each task, and all original authors choose one between $1\times10^{-5}$ and $5\times10^{-5}$. All the other settings are the same as those used in the original papers.

\section{Results and Discussion}

This section investigates how pre-trained Transformers and other pre-trained language models behave on \totallyunseensentences{} (Sec. \ref{sec:final_investigation}). Yet, to exclude that the nature of the task biases our study, we have performed additional analyses: 1) to understand whether the onion language is really different from the surface web language (Sec. \ref{sec:onions_vs_surface}), and 2) to determine the nature of the proposed classification task (Sec. \ref{sec:is_stylistic}).

\begin{table*}[h!]
\centering 
\tabcolsep=0.06cm
\begin{tabular}{l||l|c|cc|c}
\hline
& & \textbf{eBay/Legal Drugs} & \textbf{Drugs} & \textbf{Forums} & \textbf{Drugs}$\rightarrow$\textbf{Forums} \\ 

\hline

\multirow{6}{*}{\begin{sideways}\textbf{\textit{Freeze}}\end{sideways}}

&$BERT$ & $66.62(\pm 3.1)$ & $64.35(\pm 2.7)$ & $53.12(\pm 1.2)$ & $49.2(\pm 1.8)$ \\

&$Electra$ & $71.36(\pm 2.9)$ & $61.56(\pm 3.1)$ & $54.22(\pm 1.9)$ & $50.33(\pm 2.2)$ \\

&$XLNet$ & $59.92(\pm 3.4)$ & $56.77(\pm 2.7)$ & $50.32(\pm 2.6)$ & $51.86(\pm 1.8)$ \\

&$Ernie$ & $74.56(\pm 2.4)$ & $60.33(\pm 1.9)$ & $60.41(\pm 3.2)$ & $50.33(\pm 2.3)$ \\

\cline{2-6}

&\textit{BoE(GloVe)} & $91.50(\pm 0.5)$ & $81.60(\pm 1.4)$ & $54.60(\pm 1.4)$ & $53.50(\pm 1.5)$ \\ 
&\textit{KERMIT} & $90.50(\pm 1.0)$ & $79.00(\pm 1.0)$  & $66.60(\pm 1.4)$ & $\textbf{58.37}(\pm 1.26)$  \\ 
&\textit{BoE(GloVe) + KERMIT} & $93.54(\pm 1.46)$ & $83.10(\pm 1.4)$ & $66.20(\pm 1.4)$ & $54.30(\pm 2.34)$ \\ 
 
\hline
\hline
\multirow{6}{*}{\begin{sideways}\textbf{\textit{Fine-Tuning}}\end{sideways}}
&$Electra$ & $93.47(\pm 2.5)$ & $85.15(\pm 1.33)$ & $64.22(\pm 1.38)$ & $48.96(\pm 2.22)$ \\ 
&$XLNet$ & $92.58(\pm 1.99)$ & $84.95(\pm 2.13)$ & $65.32(\pm 1.69)$ & $48.22(\pm 2.46)$ \\ 
&$Ernie$ & $94.97(\pm 2.11)$ & $85.19(\pm 2.32)$ & $66.43(\pm 1.79)$ & $50.12(\pm 2.23)$ \\ 
&$BERT$ & $94.36(\pm 3.20)$ & $84.35(\pm 3.16)$ & $65.18(\pm 2.28)$ & $50.68(\pm 2.49)$ \\ 
\cline{2-6}
&$BERT_{\text{with }DomA}$ & $95.43(\pm 2.17)$ & $83.76(\pm 1.70)$ & $70.95(\pm 2.56)$ & $51.7(\pm 2.23)$ \\ 
\cline{2-6}
\cline{2-6}
&$BERT_{\text{with }ExtremeDomA}$ & $\textbf{97.4}(\pm 2.30)$ & $\textbf{89.7}(\pm 3.10)$ & $\textbf{72.4}(\pm 3.30)$ & $55.6(\pm 2.90)$ \\ 
\hline

\end{tabular}
\caption{Accuracies of pre-trained models on the Legal vs. Illegal Classification Task on the DarkWeb Corpus \cite{Choshen2019TheLO}. BERT, XLNet, Ernie, and Electra are those specific for SequenceClassification \cite{hugging}. Where applicable, models are: (1) without or with fine-tuning; (2) with domain-adaptation using only the training sets ($DomA$); (3) and, with extreme domain-adaptation using training and testing sets ($ExtremeDamA$). Experiments are obtained over 5 runs over the 5 different splits with 5 different seeds for initializing weights of neural networks.}
\label{tab:ResultsChos}
\end{table*}

\subsection{Onion Language and Surface Web Language}
\label{sec:onions_vs_surface}

\begin{figure*}[h!]
\centering
\begin{subfigure}{0.47\textwidth}
 \includegraphics[width=\linewidth]{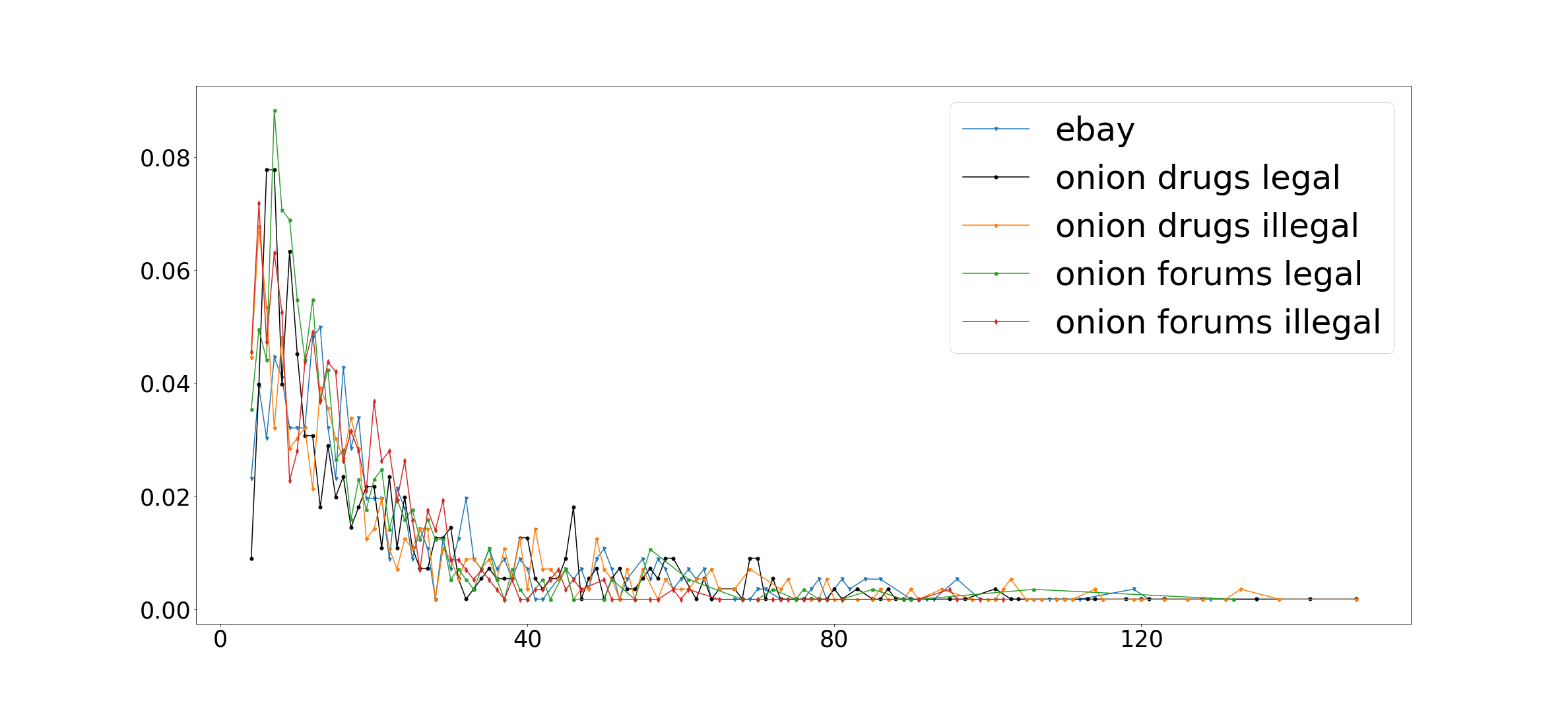}
\caption{Distribution of text length in tokens }
\label{fig:corpus_description_token}
\end{subfigure}
\hspace{5mm}
\begin{subfigure}{.47\textwidth}
 \includegraphics[width=\linewidth]{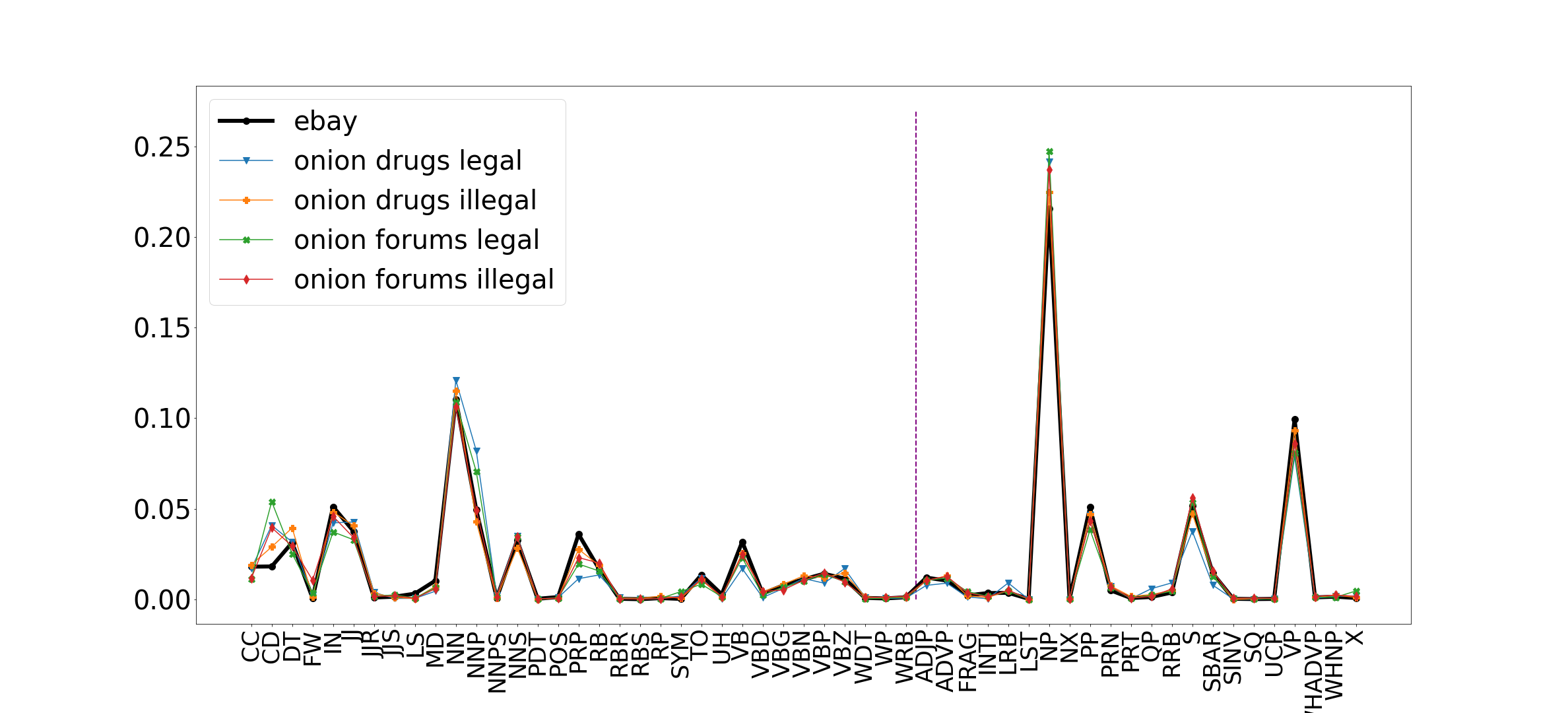}
\caption{Syntactic Analysis: POS and Non-Terminal Distribution}
\label{fig:corpus_description_labels}
 
\end{subfigure}
\caption{Corpora Facts: Analysis of the characteristics of the target corpus on the surface web and on the onion web. Syntactic analysis has been obtained by using CoreNLP \cite{zhu-etal-2013-parser-fast}}
\label{fig:corpus_description}
\end{figure*}

One important concern is whether Onion texts have some specific features that make it hard to analyze with \emph{surface-web-oriented} language analyzers. For this purpose, we compared the onion subsets with the surface web subset, that is, the \emph{eBay drugs} set (see Table \ref{tab:covered_token} and Figure \ref{fig:corpus_description}). 

Surface web and onion web texts seem to have a language with similar basic features. For example, type/token ratio of the different onion subsets is similar to the one of \emph{eBay drugs} (Table \ref{tab:covered_token}). Indeed, all these datasets contain many unique tokens as different drugs have different names (see Table \ref{tab:examples}). Moreover, the frequency of tokens vs. their length is quite similar in the surface web and the onion web (see Figure \ref{fig:corpus_description_token}). There are no strange peaks or outliers. Finally, when analyzed with a symbolic parser \cite{zhu-etal-2013-parser-fast}, these subsets have the same syntactic characteristics.
In fact, frequencies of POS tags and constituent types are similar across dataset subsets (see Figure \ref{fig:corpus_description_labels}). Indicators of not analyzed sentences such as FRAG have a similar distribution. Also, the indicators of unknown words, which are NNP and NNPS, are similar across all the subsets. 
The only differences are tags for pronouns (PRP) and cardinal numbers (CD). It seems that pronouns are more frequent for legal activities on the onion web.
Overall, onion and surface web data are mostly parsed in a similar manner.

In conclusion, the difference in the performance of the different models on the different datasets is not due to an inherent difference in the languages of the onion and surface web domains. Moreover, syntactic-based neural networks are not taking advantage of some hidden syntactic bias. 

\subsection{Illegal vs. Legal is only a Stylistic Task?}
\label{sec:is_stylistic}

Determining whether or not a piece of text is illegal seems to be a stylistic task. The hypothesis is that illegal texts are written in a way that is \emph{stylistically} different. Indeed, experiments of \citet{Choshen2019TheLO} seem to suggest that this is the case. Simple models based on POS tags using SVM or Naive Bayes have very high performances (lines 1 and 2 in Table \ref{tab:style}).

According to our experiments, the task is not only stylistic. We repeated the measures on the 5 splits we proposed (see Section \ref{sec:duta_adaptation}) and results are quite different with respect to those of \citet{Choshen2019TheLO} on a single split (lines 3 and 4 vs. lines 1 and 2 in Table \ref{tab:style}). Two datasets of eBay/Legal Drugs and Drugs seem to be strongly correlated with style and simple features. 
Yet, no strong POS tag features emerged from the Naive Bayes models and, thus, there are no apparent artifacts in these datasets. 
Instead, Forum and Drugs/Forums datasets are less correlated, if not unrelated.

Moreover, our tests with \emph{Bleaching text} model confirm that the proposed task is not solely a stylistic task. This model has been designed to capture only stylistic features \cite{van-der-goot-etal-2018-bleaching} (see Section \ref{sec:stylistic_classifiers}). Its results are quite high for eBay/Legal Drugs and Drugs and relatively low for Forum and Drugs/Forums. This is in line with the shown findings on POS-tag-based models.

\subsection{Investigating pre-trained Transformers, Lexical, and Syntactic Models}
\label{sec:final_investigation}

We can now focus on the performance of the different pre-trained models on the novel, unexplored task -- classifying legal and illegal texts in the onion web -- taking into account that there is not a real difference between the language of onion and the one of the surface web, but it is very unlikely that these \totallyunseensentences{} of the onion corpus, or very similar sentences, have been used for pre-training models.
Lexical-based and syntactic-based neural networks outperform Holistic transformers when all these models are considered universal linguistic knowledge embedders and general parameters are not fine-tuned (\emph{Freeze} in Tab. \ref{tab:ResultsChos}). Indeed, the accuracy of BoE(Glove), KERMIT, and their combination are well above the results of Holistic Transformers. It is common knowledge that Transformers need fine-tuning. However, this is the fairest comparison with other models, which cannot benefit from fine-tuning. Syntactic parsers aim to capture the general structure of language and cannot and should not be adapted to a particular task. In fact, language and knowledge about language are general, hence it is not clear why this general knowledge should be adapted to tasks with fine-tuning.  

Fine-tuning boosts the performance of holistic transformers  except for the task \emph{Drugs}$\rightarrow$\emph{Forums} (see Tab. \ref{tab:ResultsChos}). In fact, performance for the other three tasks \textit{eBay/Legal Drugs}, \textit{Drugs}, and \textit{Forums} have a dramatic increase in accuracy. To obtain these results, all layers of transformers should be fined-tuned (see Tab. \ref{tab:ResultsBERT}). Apparently, there is not a predominant set of layers that help performance to have a big increase in accuracy suggesting that some kind of linguistic knowledge is more important than another. The absence of an increase in performance for the task \emph{Drugs}$\rightarrow$\emph{Forums} is extremely interesting. Indeed, this task asks to learn a legal/illegal classifier in an environment and apply it in another environment. Fine-tuning is definitely not helping for this out-of-domain task.

Fine-tuned holistic transformers do not have an important increase in performance with respect to lexical-based and syntactic-based neural networks on these datasets with \totallyunseensentences{}. BoE(GLove)+KERMIT, which cannot be fine-tuned, are basically on par with fine-tuned transformers for the three tasks. This suggests that fine-tuning is not helping transformers to grab additional knowledge on \totallyunseensentences{}, but it seems to adapt weights to solve final tasks better. Moreover, BoE(GLove)+KERMIT and KERMIT alone still outperform transformers on the out-of-domain task \emph{Drugs}$\rightarrow$\emph{Forums}.

Extreme domain adaptation produces a real change in the performance of transformers with respect to lexical-based and syntactic-based neural networks (see last line of Tab. \ref{tab:ResultsChos}). Classical domain adaptation is not improving for \emph{Drugs} and it is improving only a little for \emph{eBay/Legal Drugs} and  \emph{Drugs}$\rightarrow$\emph{Forums}. Hence, when transformers see \totallyunseensentences{} with MLM, they seem to incorporate the knowledge needed to treat sentences in final tasks better. This may suggest that, in other classical tasks, pre-training plays a crucial role as sentences may at least have been partially seen during pre-training.

Despite all the domain adaptation and fine-tuning, holistic transformers are not gaining real clues on the difference between legal and illegal language. The best accuracy in the out-of-domain task \emph{Drugs}$\rightarrow$\emph{Forums} remains that of KERMIT, the syntax-based neural network. Hence, the compelling question is: what are these models really learning?

\subsection{Qualitative analysis}

Transformers confirm to have astonishing results if considered in a task within a single dataset and if they have partially seen sentences, that is, \emph{BERT}$_{\text{with }ExtremeDomA}$. Then, the compelling question of what they are learning should at least be addressed. For this reason, we performed a qualitative analysis of a very small part of the dataset. Focusing on examples with the most frequent range of lengths (see Fig. \ref{fig:corpus_description_token}), we selected 12 examples to better analyze the results (see Tab. \ref{tab:qualitative_analysis}). 

\begin{table}
\begin{tabular}{c|ccccc}
& \emph{oracle}& \emph{a} & \emph{m} & \emph{e} & \emph{l}\\
\hline
\emph{oracle}& - & 0.67& 0.38& 0.17& 0.17\\
\cline{3-6}
\emph{a}& 0.67\arrvline& - & 0.23& 0.17& 0.50\\
\emph{m}& 0.38\arrvline& 0.23& - & -0.04& -0.19\\
\emph{e}& 0.17\arrvline& 0.17& -0.04& - & 0.08\\
\emph{l}& 0.17\arrvline& 0.50& -0.19& 0.08& - \\
\end{tabular}
\begin{tabular}{lc}
\hline
\hline
Interannotator agreement (multi-kappa):& 0.12\\
\hline
\end{tabular}
\caption{Inter-annotation Kappa agreement matrix and multi-fleiss Kappa on a small sample of the dataset}
\label{tab:interannotationagreement}
\end{table}

The task of deciding if a text is \emph{legal} or \emph{illegal} is not simple for humans. Indeed, we asked 4 annotators to perform the task of reading the text of the examples and emitting one of the two classes. The multi-Fleiss Kappa inter-annotator agreement among these 4 annotators is very low (0.12 in Tab. \ref{tab:interannotationagreement}), which represents a slight agreement. Moreover, the majority of annotators have an agreement smaller than fair among them (see Tab. \ref{tab:interannotationagreement}). Only, one annotator ($a$) has a substantial agreement with the oracle. 
It is really difficult to decide that \emph{``All prices are in Australian dollars. (AUD) Weight: 3.5g 20 Clear''} is illegal whereas \emph{``All Major Credit, Debit, Gift, and Prepaid Cards Accepted''} is legal. Moreover, also lexical items are not really a clue. Indeed, \emph{Clomid} is both legal and illegal (see Tab. \ref{tab:qualitative_analysis}).

Performance of \emph{BERT}$_{\text{with }ExtremeDomA}$ can be then considered super-human. In three different runs with three different seeds, the BERT-based classifier has only 2 errors (last line of Tab. \ref{tab:qualitative_analysis} for runs 1 and 3). The real question is how can it be so correct for example like  \emph{``All times are UTC''} or \emph{``Do you have a coupon code?''}. During the extreme domain adaptation, BERT observes examples with Masked Language Model but it never trains on the classification task. Hence, it apparently captures handles of texts that can then be used to attach final classes.

The ability to perfectly learn in-domain classification tasks may be also the reason why \emph{BERT}$_{\text{with }ExtremeDomA}$ performs poorly in the out-of-domain task (\emph{Drugs}$\rightarrow$\emph{Forums}).

\section{Conclusions}\label{sec:DiscussConclus}

Transformers are successful in many downstream tasks, and this success also stems from the huge corpora that they are trained on. Since they are so successful, the investigation of their strengths and potential weaknesses is important. 

Our paper and our experiments show that transformers largely outperform other models only when they are pre-trained on texts which are extremely similar to texts in the target application. Indeed, only when transformers are trained with Masked Language Model (MLM) on the \totallyunseensentences{} of the DarkWeb corpus, do these transformers start to behave extremely better than other techniques. The reason why it is happening is still unclear as, in adapting to the new domain with MLM, transformers are not learning anything about the specific task but they are gaining some general model of novel texts.

Our results suggest that pre-trained transformers should clearly release the pre-training datasets to allow practitioners to explore if sentences in their dataset are included or partially covered.

In our opinion, future work should go in two directions: (1) exploring what transformers are really learning during the Masked Language Model and Next Sentence Prediction; (2) providing measures for understanding how much a pre-trained model knows about given texts and given datasets as started in \cite{ranaldi-etal-2023-precog}.

\newpage
\section*{Acknoledgements}
This paper has been supported by “Knowledge At the Tip of Your fingers: Clinical Knowledge for Humanity” (KATY) project funded by the European Union’s Horizon 2020 research and innovation program under grant agreement No. 101017453.

We would also like to thank all the anonymous reviewers who helped to strengthen this paper. 

\section*{Limitations}

We believe that the main results obtained in this paper are convincing: Transformers behave better if they see in advance, at least, part of the corpus. Yet, our paper leaves some open avenues to explore besides the two future research lines described in the conclusions.

One limitation is due to the fact that we explored only one possible corpus with \totallyunseensentences{}. To assess these results better, additional corpora should be taken into consideration. For researchers outside big companies, retrieving such corpora is extremely difficult. As a possible solution, these corpora should be retrieved where they naturally and publicly occur. 

\section*{Ethics Statement}

Navigating the Dark Web may be extremely dangerous. Indeed, it may contain offensive content or illegal content, and tasks themselves could potentially have harmful content. We really need to describe how this is navigated and, more importantly, what are the real pros of using such resources. 

We propose to navigate the Dark Web in a textual manner so that there is no exposure and no need to download sensitive visual material that may result in a crime. In this textual version of the Dark Web, systems may be exposed to offensive or illegal content, but offensive content is everywhere, and then it is not a real problem. 

However, using Dark Web material is a way to access really unseen text, which has never been used in Pre-trained Transformers. Unlike what we believe can happen in large companies, public researchers hardly have the possibility to access user-produced data that are not seen. This is a democratic way to obtain such truly unseen data.

\bibliographystyle{acl_natbib}
\bibliography{my_bib}

\newpage
\appendix
\begin{table*}[t]
\section{Appendix}

\centering
\begin{small}
\begin{tabular}{p{5cm}|p{5cm}|p{5cm}}
\multicolumn{1}{c}{\textbf{eBay drugs}}                     & \multicolumn{1}{c}{\textbf{Legal Onion}}                     & \multicolumn{1}{c}{\textbf{Illegal Onion}}                   \\ \hline

Asclepias (butterfly weed) seeds. This E-Z grower is a butterfly magnet \& host plant for the Queen \& Monarch butterfly. Loves full sun, is drought tolerant and prefers sandy, dry soil or gravel soil. It is a favorite of hummingbirds. 20 seeds per package.
& Generic Synthroid is used for treating low thyroid activity and treating or suppressing different types of goiters. It is also used with surgery and other medicines for managing certain types of thyroid cancer. & 
Known as: Clomid / Clofi / Fertomid / Milophene / Wellfert Generic Clomid is used for treating female infertility. Select dosage 100mg Package Price Per pill Savings Order 25mg x 30 pills Price: \$29.95 Per pill: \$1.00 \\ \hline

Big book-opening drugs bag 2 parts with adjustable elastics to feature 60 different sized ampoules Central padding with transparent pocket for ampoules list and expiry date. Made in red, tear-resistant, water-resistant. & Propecia is used for treating certain types of male pattern hair loss (androgenic alopecia) in men. It is also used to treat symptoms of benign prostatic hyperplasia (BPH) in men with an enlarged prostate. & TAMOXIFEN blocks the effects of estrogen. It is commonly used to treat breast cancer. It is also used to decrease the chance of breast cancer coming back in women who have received treatment for the disease.  \\ \hline
\end{tabular}
\end{small}
\caption{Example paragraphs (data instances) taken from the Legal Onion and Illegal Onion subsets training sets of the drug-related corpus. Each paragraph is reduced to the first 50 characters for space reasons.}
\label{tab:examples}

\section{Appendix}

\centering 
\begin{small}
 \begin{tabular}{lc|cccccc}
               &             & \multicolumn{2}{c}{BERT} & \textit{Electra} &  \textit{XLNet} & \textit{Ernie} & KERMIT\\
 \emph{Corpus} & \emph{Size} &$_{base}$ &  $_{multi}$ & &&& (Parser)\\ 
 \hline
BooksCorpus \cite{zhu2015aligning}   & 800M words   & $\surd$ &  & $\surd$ & $\surd$ &  \\ 
2010-and-2014-English Wikipedia dump & 2,500M words & $\surd$ & $\surd$ & $\surd$ & $\surd$ & $\surd$ & \\
Giga5 \cite{parker2011english}        & 16GB     & $\surd$ &  & $\surd$ & $\surd$ &  &\\
Common Crawl \cite{commoncrawl}      & 110GB    &  & & & $\surd$ & &\\
ClueWeb \cite{callan2009clueweb09}   & 19GB    &  & & & $\surd$ & &\\
Penn Treebank \cite{marcus-etal-1993-building} & 1M words &  & & & & & $\surd$\\ 
 \end{tabular}
 \end{small}
 \caption{Pre-traning corpora with their size. All corpora are derived from the surface web.}
 \label{tab:pretraining}
\end{table*}

\newpage

\begin{table*}[t]
\section{Appendix}

\begin{small}
\begin{tabular}{p{10cm}ccc}
\emph{Text}	&	\emph{Oracle}  & \multicolumn{2}{c}{\emph{BERT}$_{\text{with }ExtremeDomA}$} \\
            &	                & Runs 1 and 3 & Run 2 \\
\hline
You should never take more than one dose more than once a day.	&	illegal &illegal &	illegal\\
4. Fill in the order information required	&	illegal &illegal &	illegal\\
All Items Ship Via First Class Air Mail - registered or unregistered, add an extra \$25 and we ship express mail EMS	&	legal &	legal& legal\\
All Major Credit, Debit, Gift, and Prepaid Cards Accepted	&	legal&	legal&legal\\
All prices are in Australian dollars. (AUD) Weight: 3.5g 20 Clear	&	illegal&	illegal &illegal\\
All times are UTC	&	illegal&	illegal &illegal\\
aunice September 15, 2016 Super fast shipping. Great product as always.	&	illegal&	illegal &illegal\\
Balkan Pharmaceuticals Ltd. (Moldova) Turinabol	&	legal&	legal&legal\\
Do you have a coupon code?	&	legal&	legal&legal\\
free shipping On all orders \$50.00 or more	&	legal&	legal&legal\\
Generic Clomid is used for treating female infertility. More Info	&	legal&	legal&legal\\
Known as: Clomid / Clofi / Fertomid / Milophene / Ovamid / Serophene / Wellfert Generic Clomid is used for treating female infertility. Select dosage 100mg Package Price Per pill Savings Order 25mg x 30 pills Price:\$ 29.95 Per pill:\$ 1.00 Order:	&	illegal&	legal & illegal\\
\hline
\end{tabular}
\end{small}
\caption{Dataset Drugs: Oracle classifications along with classifications of three runs of \emph{BERT}$_{\text{with }ExtremeDomA}$ with three different seeds.}
\label{tab:qualitative_analysis}

\section{Appendix}
\centering 
\tabcolsep=0.06cm
\begin{tabular}{llcc|cccc}
\hline
& Model & &   & \textbf{eBay/Legal Drugs} & \textbf{Drugs} & \textbf{Forums} & \textbf{Drugs/Forums} \\ 

\hline

&$BERT$ && & $94.36(\pm 3.20)$ & $84.35(\pm 3.16)$ & $65.18(\pm 2.28)$ & $50.68(\pm 2.49)$ \\ 

\hline

&$BERT_{last\text{ }2}$ && & $73.25(\pm 2.6)$ & $68.26(\pm 3.4)$ & $59.94(\pm 2.8)$ & $49.93(\pm 3.6)$ \\ 

&$BERT_{last\text{ }4}$ && & $80.02(\pm 2.2)$ & $70.62(\pm 2.8)$ & $59.11(\pm 2.9)$ & $50.75(\pm 1.9)$ \\ 

&$BERT_{last\text{ }6}$ && & $86.89(\pm 2.9)$ & $71.47(\pm 2.6)$ & $60.56(\pm 1.9)$ & $52.17(\pm 2.9)$ \\ 

&$BERT_{last\text{ }8}$ && & $77.62(\pm 2.6)$ & $75.22(\pm 3.2)$ & $65.08(\pm 2.7)$ & $50.81(\pm 2.4)$ \\ 

&$BERT_{last\text{ }10}$ && & $89.95(\pm 2.4)$ & $79.37(\pm 2.7)$ & $62.96(\pm 4.2)$ & $50.11(\pm 2.7)$ \\

\hline

&$BERT_{\text{with }DomA}$  & && $95.43(\pm 2.17)$ & $83.76(\pm 1.70)$ & $70.95(\pm 2.56)$ & $51.7(\pm 2.23)$ \\ 

&$BERT_{\text{with }ExtremeDomA}$ & && $\textbf{97.4}(\pm 2.30)$ & $\textbf{89.7}(\pm 3.10)$ & $\textbf{72.4}(\pm 3.30)$ & $\textbf{55.6}(\pm 2.90)$ \\

\hline

\end{tabular}
\caption{Accuracies for $BERT_{base}$: (1) fine-tuned on the $last\text{ }n$ layers; (2) domain-adapted (DomA) without and with fine-tuning; (3) sentence-adapted (SenA)  without and with fine-tuning. Experiments are obtained over 5 runs with different seeds. }
\vspace{-1.7em}
\label{tab:ResultsBERT}

\end{table*}

\end{document}